\pdfoutput=1

\documentclass[11pt]{article}

\usepackage[]{acl}

\usepackage{times}
\usepackage{latexsym}

\usepackage{graphicx}
\usepackage{amsmath}
\usepackage{booktabs}
\usepackage{cleveref}
\usepackage{multirow}
\usepackage{hyperref}
\usepackage{amssymb}
\usepackage{pifont}
\usepackage{color, colortbl}
\usepackage{comment}
\usepackage{subcaption}

\usepackage[T1]{fontenc}

\usepackage[utf8]{inputenc}

\usepackage{microtype}
\usepackage{bm}

\usepackage[utf8]{inputenc}

\def\eqref#1{equation~\ref{#1}}

\def\1{\bm{1}}

\def\vh{{\bm{h}}}

\def\vs{{\bm{s}}}

\def\vv{{\bm{v}}}

\DeclareMathAlphabet{\mathsfit}{\encodingdefault}{\sfdefault}{m}{sl}
\SetMathAlphabet{\mathsfit}{bold}{\encodingdefault}{\sfdefault}{bx}{n}

\definecolor{Color}{gray}{0.9}

\title{MCSE: Multimodal Contrastive Learning of Sentence Embeddings} 

\author{ Miaoran Zhang, Marius Mosbach, David Ifeoluwa Adelani, \\ \textbf{Michael A. Hedderich,} and \textbf{Dietrich Klakow} \\
  Spoken Language Systems (LSV) \\
  Saarland Informatics Campus, Saarland University, Germany \\
  \texttt{\{mzhang,mmosbach,didelani\}@lsv.uni-saarland.de} \\
  \texttt{\{mhedderich,dklakow\}@lsv.uni-saarland.de} \\}

\begin{document}
\maketitle

\begin{abstract}
Learning semantically meaningful sentence embeddings is an open problem in natural language processing. In this work, we propose a sentence embedding learning approach that exploits both visual and textual information via a multimodal contrastive objective. Through experiments on a variety of semantic textual similarity tasks, we demonstrate that our approach consistently improves the performance across various datasets and pre-trained encoders. In particular, combining a small amount of multimodal data with a large text-only corpus, we improve the state-of-the-art average Spearman's correlation by $1.7\%$. By analyzing the properties of the textual embedding space, we show that our model excels in aligning semantically similar sentences, providing an explanation for its improved performance.

\end{abstract}

\section{Introduction}
Sentence embedding learning, i.e., encoding sentences into fixed-length vectors that faithfully reflect the semantic relatedness among sentences, is a fundamental challenge in natural language processing (NLP). Despite the tremendous success of pre-trained language models (PLMs), such as BERT~\citep{devlin2019bert} and RoBERTa~\citep{liu2019roberta}, it has been shown that the off-the-shelf sentence embeddings of PLMs without fine-tuning are even inferior to averaging Glove embeddings~\citep{pennington2014glove} in terms of semantic similarity measure  ~\citep{reimers-gurevych-2019-sentence}. Hence, recent research~\citep{li-etal-2020-sentence, zhang-etal-2020-unsupervised, su2021whitening} focuses on adjusting the original sentence embeddings derived from PLMs in an unsupervised manner. In particular, there has been growing interest in adopting contrastive learning objectives to achieve this goal~\citep{carlsson2020semantic, kim-etal-2021-self, gao2021simcse}.

Although purely text-based models have led to impressive progress, it remains an open question to what extent they capture the deeper notion of sentence meaning beyond the statistical distribution of texts, which lies outside of the text and is grounded in the real-world ~\citep{bender-koller-2020-climbing, bisk-etal-2020-experience}. As a central part of the human perceptual experience, vision has been shown to be effective in grounding language models and improving performance on various NLP tasks~\citep{zhang2019neural, bordes2019incorporating, zhao2020visually}. We hypothesize that using vision as supplementary semantic information can further promote sentence representation learning.

In this work, we propose \textit{MCSE}, an approach for \underline{m}ultimodal \underline{c}ontrastive learning of \underline{s}entence \underline{e}mbeddings. To exploit both visual and textual information, we adopt the state-of-the-art contrastive sentence embedding framework SimCSE~\citep{gao2021simcse} and extend it with a multimodal contrastive objective. In addition to the textual objective in SimCSE that maximizes agreement between positive sentence pairs, the multimodal objective maximizes agreement between sentences and corresponding images in a shared space. We conduct extensive experiments on standard Semantic Textual Similarity (STS) benchmarks and show the effectiveness of MCSE across various datasets and pre-trained encoders. We find that, using a small amount of multimodal data in addition to a text-only corpus yields significant improvements on STS tasks. By analyzing the alignment and uniformity properties of the embedding space~\citep{wang2020understanding}, we show that MCSE better aligns the semantically similar sentences while maintaining uniformity, providing an explanation for its superior performance.\footnote{Our code and pre-trained models are publicly available at \href{https://github.com/uds-lsv/MCSE}{https://github.com/uds-lsv/MCSE}.}

\begin{figure*}[t]
    \centering
    \includegraphics[scale=0.72]{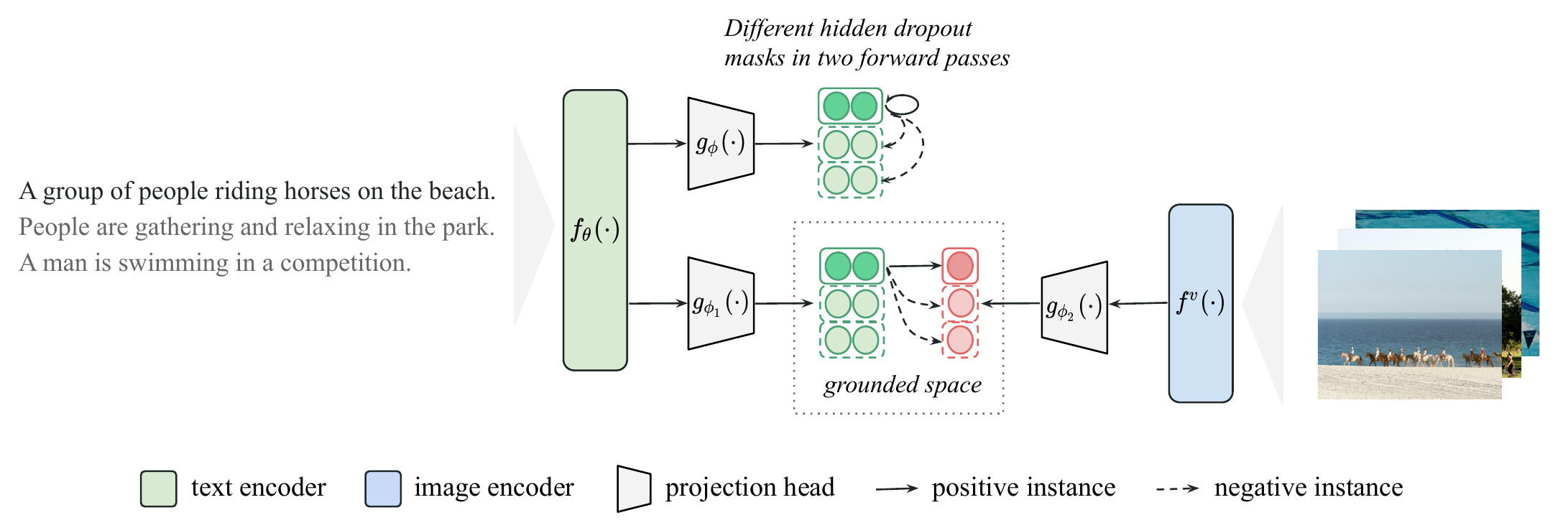} 
    \caption{The overall architecture of MCSE. Compared to SimCSE,
    a new multimodal objective is calculated in the grounded space. For each input sentence, the positive instance is the paired image and the negative instances are all other in-batch images. }
    \vspace{-2mm}
    \label{fig:arch}
\end{figure*}

\section{Related Work}
\textbf{Sentence Representation Learning.} Existing works for learning sentence embeddings can be categorized into supervised~\citep{conneau2017supervised, cer-etal-2018-universal, reimers-gurevych-2019-sentence, wieting2020bilingual} and unsupervised approaches~\citep{li-etal-2020-sentence, carlsson2020semantic, su2021whitening, kim-etal-2021-self, gao2021simcse, liu-etal-2021-fast, yan-etal-2021-consert}. Supervised approaches mostly utilize supervision from annotated natural language inference data or parallel data. Unsupervised approaches are able to make use of the intrinsic semantic information embedded in the natural language text corpus by adjusting the training objective to STS tasks, thereby eliminating the need for a costly annotation process. In particular, contrastive learning objective~\citep{carlsson2020semantic, kim-etal-2021-self, gao2021simcse, liu-etal-2021-fast, yan-etal-2021-consert} regularizes the embedding space by pulling positive (i.e., semantically similar) sentences closer and pushing apart negatives, showcasing great effectiveness in capturing the semantic similarity among sentences. Our approach adopts the contrastive learning framework and is built on top of the current state-of-the-art approach \citep{gao2021simcse}, further pushing the frontier of STS by leveraging multimodal semantic information.

\noindent
\textbf{Visually Grounded Representation Learning.}
There are various works showing that grounding NLP models to the visual world can improve textual representation learning. %
\citet{lazaridou2015combining} and \citet{zablocki2018learning} learn word embeddings by aligning words to the visual entity or visual context. \citet{kiela2018learning} ground sentence embeddings by predicting both images and alternative captions related to the same image.  \citet{bordes2019incorporating} enhance the Skip-Thought model~\citep{kiros2015skip} by learning a grounded space that preserves the structure of visual and textual spaces. 
Recently, \citet{tan2020vokenization} and \citet{tang2021vidlankd} train large scale language models with multimodal supervision from scratch with the goal of improving general language understanding. Different from the aforementioned works, we focus on learning visually grounded sentence embeddings by fine-tuning pre-trained models in a contrastive learning framework.

\begin{table*}[t]
 \begin{center}
 \scalebox{0.75}{
  \begin{tabular}{clccccccc|c}
    \toprule
    &\textbf{Model} & \textbf{STS12} & \textbf{STS13} & \textbf{STS14} & \textbf{STS15} & \textbf{STS16} & \textbf{STS-B} & \textbf{SICK-R} & \textbf{Avg.$\uparrow$} \\
    \midrule
    \midrule
    & BERT (first-last avg.) & 39.7 & 59.4 & 49.7 & 66.0 & 66.2 & 53.9 & 62.1 & 56.7\\ 
    &RoBERTa (first-last avg.) & 40.9 & 58.7 &	49.1 &	65.6&	61.5 &	58.6 &	61.6 &	56.6 \\
    \midrule
    \parbox[t]{2mm}{\multirow{4}{*}{\rotatebox[origin=c]{90}{\textit{wiki}}}}
    &SimCSE-BERT$^\diamondsuit$ & 68.4& 82.4& 74.4& 80.9& 78.6& 76.9& 72.2& 76.3\\
    &SimCSE-RoBERTa$^\diamondsuit$  & 70.2& 81.8& 73.2& 81.4& 80.7& 80.2& 68.6& 76.6\\
    \cmidrule{2-10}
    &SimCSE-BERT & 
    67.8$_{\pm 1.6}$ &
    80.0$_{\pm 2.1}$ &
    72.5$_{\pm 1.7}$ &
    80.1$_{\pm 0.8}$ &
    77.6$_{\pm 0.8}$ &
    76.5$_{\pm 0.8}$ &
    70.1$_{\pm 0.9}$ & 
    74.9$_{\pm 1.1}$ \\ 
    &SimCSE-RoBERTa &
    68.7$_{\pm1.0}$ &
    82.0$_{\pm0.5}$ &
    74.0$_{\pm1.0}$ &
    82.1$_{\pm0.4}$ &
    81.1$_{\pm0.4}$ &
    80.6$_{\pm0.3}$ &
    69.2$_{\pm0.2}$ &
    76.8$_{\pm0.5}$\\
    \midrule
    \parbox[t]{2mm}{\multirow{4}{*}{\rotatebox[origin=c]{90}{\textit{wiki+flickr}}}}  &SimCSE-BERT & 
    69.9$_{\pm 1.7}$ &
    79.8$_{\pm 1.5}$ &
    72.9$_{\pm 0.9}$ &
    81.9$_{\pm 0.8}$ &
    \textbf{77.8}$_{\pm 0.9}$ &
    76.6$_{\pm 1.1}$ &
    68.4$_{\pm 0.8}$ &
    75.3$_{\pm 0.9}$  \\ 
    & \cellcolor{Color}MCSE-BERT & 
    \cellcolor{Color}\textbf{71.4}$_{\pm 0.9}$ &
    \cellcolor{Color} \textbf{81.8}$^*_{\pm 1.3}$ &
    \cellcolor{Color} \textbf{74.8}$^*_{\pm 0.9}$ &
     \cellcolor{Color}\textbf{83.6}$_{\pm 0.9}$ &
     \cellcolor{Color}77.5$_{\pm 0.8}$ &
     \cellcolor{Color}\textbf{79.5}$^*_{\pm 0.5}$ &
     \cellcolor{Color}\textbf{72.6}$^*_{\pm 1.4}$ &
     \cellcolor{Color}\textbf{77.3}$^*_{\pm 0.5}$  \\ 
    \cmidrule{2-10}
    &SimCSE-RoBERTa  &
    69.5$_{\pm0.9}$&
    81.6$_{\pm0.5}$&
    74.1$_{\pm0.6}$&
    82.4$_{\pm0.3}$&
    80.9$_{\pm0.5}$&
    79.9$_{\pm0.3}$&
    67.3$_{\pm0.5}$&
    76.5$_{\pm0.4}$ \\ 
    &  \cellcolor{Color}MCSE-RoBERTa  &
     \cellcolor{Color}\textbf{71.7}$^*_{\pm0.2}$ &
     \cellcolor{Color}\textbf{82.7}$^*_{\pm0.4}$&
     \cellcolor{Color}\textbf{75.9}$^*_{\pm0.3}$&
     \cellcolor{Color}\textbf{84.0}$^*_{\pm0.4}$&
     \cellcolor{Color}\textbf{81.3}$_{\pm0.3}$&
     \cellcolor{Color}\textbf{82.3}$^*_{\pm0.5}$&
     \cellcolor{Color}\textbf{70.3}$^*_{\pm1.3}$&
    \cellcolor{Color}\textbf{78.3}$^*_{\pm0.1}$ \\ 
    \midrule
    \parbox[t]{2mm}{\multirow{4}{*}{\rotatebox[origin=c]{90}{\textit{wiki+coco}}}}&SimCSE-BERT  &  
    69.1$_{\pm 1.0}$ &
    80.4$_{\pm 0.9}$ &
    72.7$_{\pm 0.7}$ &
    81.1$_{\pm 0.3}$ &
    \textbf{78.2}$_{\pm 0.9}$ &
    73.9$_{\pm 0.6}$ &
    66.6$_{\pm 1.2}$ &  
    74.6$_{\pm 0.2}$ \\ 
    & \cellcolor{Color} MCSE-BERT &
     \cellcolor{Color}\textbf{71.2}$^*_{\pm 1.3}$ &
     \cellcolor{Color}\textbf{79.7}$_{\pm 0.9}$ &
     \cellcolor{Color}\textbf{73.8}$_{\pm 0.9}$ &
     \cellcolor{Color}\textbf{83.0}$^*_{\pm 0.4}$ &
     \cellcolor{Color}77.8$_{\pm 0.9}$ &
     \cellcolor{Color}\textbf{78.5}$^*_{\pm 0.4}$ &
     \cellcolor{Color}\textbf{72.1}$^*_{\pm 1.4}$ &
     \cellcolor{Color}\textbf{76.6}$^*_{\pm 0.5}$ \\ 
    \cmidrule{2-10}
    & SimCSE-RoBERTa &
    66.4$_{\pm 0.9}$&
    80.7$_{\pm 0.7}$&
    72.7$_{\pm 1.1}$&
    81.3$_{\pm 0.9}$&
    80.2$_{\pm 0.8}$&
    76.8$_{\pm 0.6}$&
    65.7$_{\pm 0.7}$&
    74.8$_{\pm 0.5}$ \\ 
    &  \cellcolor{Color}MCSE-RoBERTa &
     \cellcolor{Color}\textbf{70.2}$^*_{\pm 1.7}$&
     \cellcolor{Color}\textbf{82.0}$^*_{\pm 0.7}$&
     \cellcolor{Color}\textbf{75.5}$^*_{\pm 1.2}$&
     \cellcolor{Color}\textbf{83.0}$^*_{\pm 0.6}$&
     \cellcolor{Color}\textbf{81.5}$^*_{\pm 0.7}$&
     \cellcolor{Color}\textbf{80.8}$^*_{\pm 1.0}$&
     \cellcolor{Color}\textbf{69.9}$^*_{\pm 0.6}$&
     \cellcolor{Color}\textbf{77.6}$^*_{\pm 0.8}$ \\
  \bottomrule
  \end{tabular}}\\
  \vspace{1mm}
      \begin{small}
      $\ast$: difference between SimCSE and MCSE is significant at $\alpha=0.05$ according to an independent t-test.
     \end{small}
  \caption{Performance comparison on STS tasks. STS-B: STS Benchmark, SICK-R: SICK-Relatedness, Avg.: average across $7$ tasks. $\diamondsuit$ : single seed results from \citet{gao2021simcse}. All other results are from our implementation. Models are trained with $5$ random seeds and we report the means and standard deviations. }
  \vspace{-2mm}
  \label{tab:wiki}
  \end{center}
\end{table*}

\section{Method}
To exploit both visual and textual information, we adopt SimCSE \citep{gao2021simcse} as the textual baseline and extend it with a multimodal contrastive learning objective. 

\subsection{Background: Unsupervised SimCSE}
\label{sec:simcse}
Data augmentation plays a critical role in contrastive self-supervised representation learning~\citep{chen2020simple}. The idea of unsupervised SimCSE is to use dropout noise as a simple yet effective data augmentation strategy. Given a collection of sentences $\{x_i\}_{i=1}^m$, we construct a positive pair for each input $x_i$ by encoding it twice using different dropout masks: $\vh_i^{z}=g_{\phi}(f_{\theta}(x_i, z))$ and $\vh_i^{z'}=g_{\phi}(f_{\theta}(x_i, z'))$, where $z$ and $z'$ denote different dropout masks\footnote{The standard dropout masks in Transformers are used.}, $f_{\theta}(\cdot)$ is a pre-trained language encoder such as BERT, and $g_{\phi}(\cdot)$ is a projection head\footnote{There is a MLP pooler layer over \texttt{[CLS]} in BERT's implementation. \citet{gao2021simcse} use it with re-initialization.} on top of the \texttt{[CLS]} token. The training objective is: %

\begin{equation}
    \ell_i^\mathrm{S}=-\log \frac{e^{\mathrm{sim}(\vh_i^{z_i}, \vh_i^{z_i'})/ \tau}}{ \sum_{j=1}^N e^{\mathrm{sim}(\vh_i^{z_i}, \vh_j^{z_j'})/ \tau}} \;,
\end{equation}

where $N$ is the size of the mini-batch, $\tau$ is a temperature parameter and $\mathrm{sim} (\vh_1, \vh_2)$ is the cosine similarity $\frac{\vh_1^T \vh_2}{ \left\|\vh_1 \right\|\cdot \left\|\vh_2 \right\|  }$. After training, the \texttt{[CLS]} token outputs of the language encoder are taken as the sentence embeddings.

\subsection{Multimodal Contrastive Learning} 
\label{sec:mcse}
Beyond the textual objective in SimCSE, we introduce a multimodal objective within the contrastive learning framework. The overview of our MCSE model is shown in Figure \ref{fig:arch}. Given a collection of sentence-image pairs $D=\{ x_i, y_i  \}_{i=1}^m$, firstly we map sentence $x_i$ and image $y_i$  into a shared space:
\begin{equation}
    \vs_i^z = g_{\phi_1}(f_{\theta}(x_i, z)),  \,
    \vv_i = g_{\phi_2} (f^v(y_i)) \; ,
\end{equation}
where $f^v(\cdot)$ is a pre-trained image encoder such as ResNet~\citep{he2016deep}, which is fixed during training. $g_{\phi_1}(\cdot)$ and $g_{\phi_2}(\cdot)$ are distinct projection heads for text and image modality respectively. To pull semantically close image-sentence pairs together and push away non-related pairs, we define the multimodal contrastive learning objective as:
\begin{equation}
    \ell_i^\mathrm{M}=- \sum_{z\in\{ z_i, z_i' \}}\log \frac{e^{\mathrm{sim}(\vs_i^z, \vv_i)/ \tau'}}{ \sum_{j=1}^N e^{\mathrm{sim}(\vs_i^z,\vv_j)/ \tau'}} \; ,
\end{equation}

where $\tau'$ is a temperature parameter. Let $\lambda$ denote the trade-off hyperparameter between two objectives, we formulate the final loss as:
\begin{equation}
\ell_i = \ell_i^S + \lambda  \ell_i^M \; .
\end{equation}
Our method further regularizes the sentence representation in a way that aligns with the image representation in the grounded space.

\section{Experiments}
\subsection{Setup}
\textbf{Dataset}
We use Flickr30k~\citep{young2014flickr} and MS-COCO~\citep{lin2014coco} as our multimodal datasets. Flickr30k contains $29,783$ training images and MS-COCO contains $82,783$ training images. Each image is annotated with multiple captions and we randomly sample only one caption to create image-sentence pairs. Following \citet{gao2021simcse}, we use  Wiki1M as the text-only corpus, which consists of $10^6$ sentences randomly drawn from English Wikipedia. 

\noindent
\textbf{Implementation Details} 
We use BERT$_{base}$~\citep{devlin2019bert} and RoBERTa$_{base}$~\citep{liu2019roberta} as language encoders and ResNet-50~\citep{he2016deep} as the image encoder. Distinct single-layer MLPs are applied as projection heads. More details are provided in Appendix \ref{sec:appendix_implementation}.

\noindent
\textbf{Evaluation} We evaluate the trained models on seven Semantic Textual Similarity (STS) tasks: STS 2012-2016~\citep{agirre2012semeval, agirre2013sem, agirre2014semeval, agirre2015semeval, agirre2016semeval}, STS Benchmark~\citep{cer2017semeval}  and SICK-Relatedness~\citep{marelli2014sick}. 
Each of these datasets consists of a collection of sentence pairs and the goal is to predict a similarity score for each sentence pair. Following \citet{gao2021simcse}, we report the Spearman's correlation ($\times 100$) between gold annotations and predicted scores in the ``all'' setting, i.e., for each task, we concatenate all the subsets and report the overall Spearman's correlation.  

\subsection{Main Results}
\paragraph{Augmenting text-only corpus with small scale multimodal data yields significant improvements.} To fully utilize different types of data resources, we conduct experiments with a text-only corpus and multimodal data. SimCSE is trained on sentences and captions only, while MCSE additionally computes the multimodal objective for image-caption pairs. As shown in Table \ref{tab:wiki}, averaging the off-the-shelf BERT and RoBERTa embeddings\footnote{Following~\citep{gao2021simcse}, we take the average of the first and last layers, which is better than only using the last.} yields poor performance on STS tasks. SimCSE models significantly outperform the average embeddings. MCSE models, which have access to auxiliary visual information, further achieve noticeable improvements even if the amount of multimodal data is relatively small. When MCSE is applied to the combination of Wiki1M and Flickr30k, it improves the state-of-the-art result for BERT ($76.3\rightarrow77.3$) and RoBERTa ($76.6\rightarrow78.3$) by a decent margin. Looking at performance on the individual tasks, we find that MCSE models using BERT encoder perform worse on STS16. This can be attributed to the domain discrepancy, where some subsets that are close to the training distribution benefit more from visually grounding than others (see Appendix \ref{sec:appendix_subset}). 

To further investigate the impact of different datasets, we train models solely on multimodal data and report results in Table \ref{tab:caption}. We observe that, without the large text-only corpus, the performances decrease considerably compared to results in Table \ref{tab:wiki}.
Still, MCSE models consistently surpass SimCSE models ($0.9$~--~$3.8$ points improvement). Moreover, replacing the paired images with shuffled images before training MCSE leads to $0.8$~--~$5.0$ points reduction in terms of average Spearman's correlation, further validating the efficacy of visual semantics. We also replace the ResNet encoder with CLIP \citep{radford2021clip} and our results show that different image encoders lead to similar results. Details are shown in Appendix \ref{sec:appendix_ablation}.  

\begin{table}[h]
 \begin{center}
 \scalebox{0.75}{
  \begin{tabular}{lcc}
    \toprule
    \multirow{2.5}{*}{\textbf{Model}} & \multicolumn{2}{c}{\textbf{Trained on}}\\
    \cmidrule{2-3}
     & \textit{flickr} & \textit{coco}\\
    \midrule
    \midrule
    SimCSE-BERT & 68.8$_{\pm0.7}$ & 67.8$_{\pm0.4}$ \\
    \rowcolor{Color} 
    MCSE-BERT & \textbf{70.6}$^*_{\pm0.5}$ & \textbf{71.6}$^*_{\pm0.2}$ \\
    \qquad w/ shuffling & 67.9$_{\pm 0.6}$\textcolor{red}{$\downarrow$} & 66.6$_{\pm 0.3}$\textcolor{red}{$\downarrow$} \\
    \midrule
    SimCSE-RoBERTa & 72.9$_{\pm0.3}$ & 72.8$_{\pm0.3}$ \\
    \rowcolor{Color} 
    MCSE-RoBERTa & \textbf{73.8}$^*_{\pm0.2}$ & \textbf{74.3}$^*_{\pm0.3}$ \\
    \qquad w/ shuffling & 73.0$_{\pm 0.4}$\textcolor{red}{$\downarrow$} & 72.8$_{\pm 0.3}$\textcolor{red}{$\downarrow$} \\
    \bottomrule
  \end{tabular}}\\
  \vspace{1mm}
      \begin{small}
      $\ast$: difference between SimCSE and MCSE is significant.
     \end{small}
  \caption{ Comparison of the average Spearman's correlation on $7$ STS tasks (Avg. column in Table \ref{tab:wiki}). We report the means and standard deviations over $5$ seeds.}
  \label{tab:caption}
  \end{center}
\end{table}

\begin{figure}[t]
    \centering
    \includegraphics[scale=0.66]{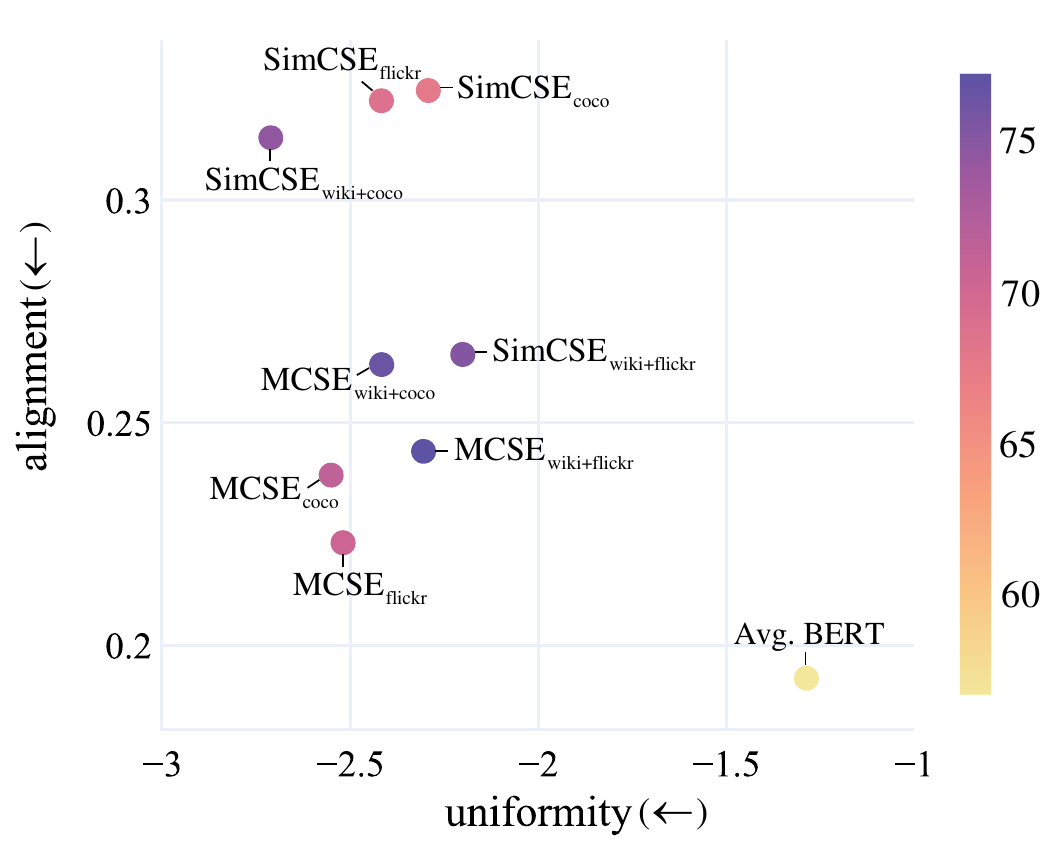} 
    \caption{The alignment-uniformity plot of models when using BERT encoder.  Colors of dots represent the average Spearman’s correlation.}
    \label{fig:metric}
\end{figure}

\begin{figure*}[ht]
    \centering
    \includegraphics[scale=0.65]{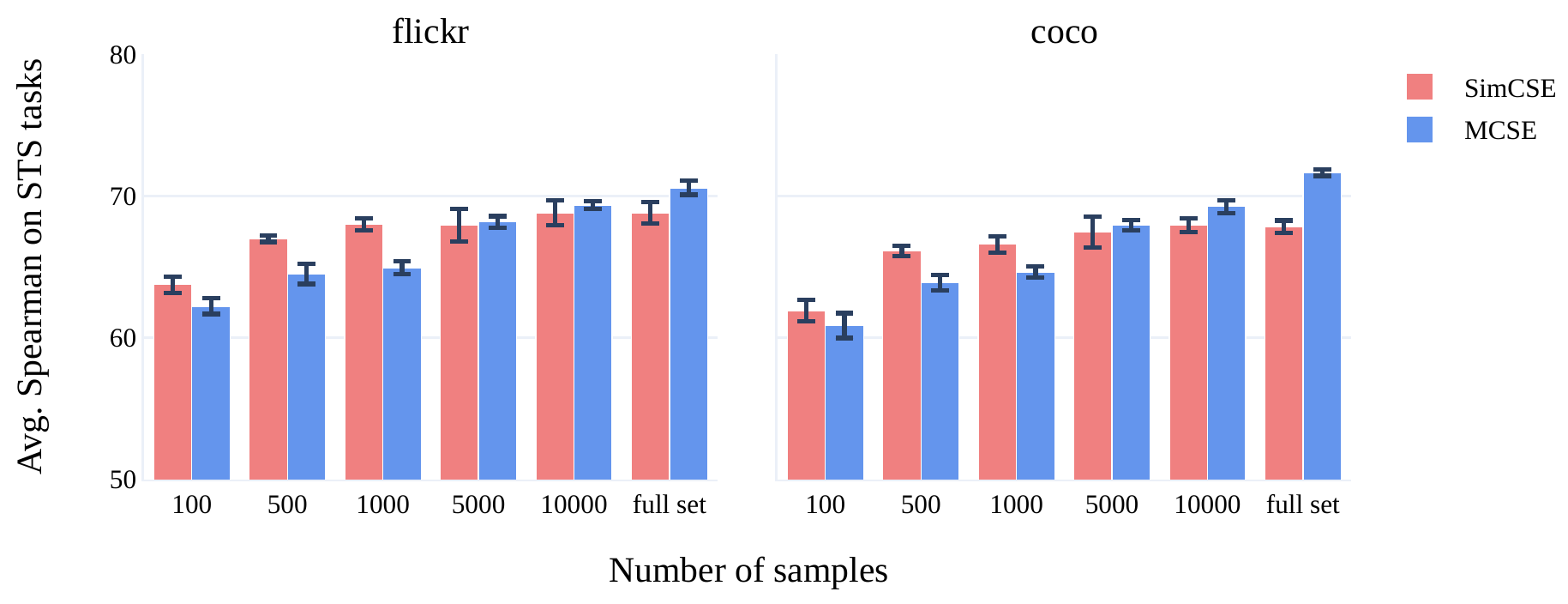} 
    \caption{Performances of different data scales. The full set indicates 30K and 87K samples for Flickr30k and MS-COCO respectively.}
    \label{fig:scale}
    \vspace{-2mm}
\end{figure*}

\paragraph{Grounding to the visual world improves alignment and maintains uniformity.}
To dissect the inner workings of MCSE, we use two quantifiable metrics proposed in \citet{wang2020understanding}: \textit{alignment} and \textit{uniformity}, as measurements of representation quality. Let $p_{\text{pos}}$ denote the positive pairs distribution and $p_{\text{data}}$ denote the data distribution. The \textit{alignment loss} prefers encoders that assign similar features to semantically similar instances (assuming features have been normalized): 
\begin{equation}
    \mathcal{L}_{align}\triangleq \mathop{\mathbb{E}}_{(x, x^+)~\sim~ p_{\text{pos}}}\left\| f(x)-f(x^+) \right\|^2_2 \; .
\end{equation}
And the \textit{uniformity loss} prefers a uniform distribution in the hypersphere:
\begin{equation}
    \mathcal{L}_{uniform}\triangleq \log \mathop{\mathbb{E}}_{x,y \stackrel{i.i.d.}{\sim} p_{\text{data}}} e^{-2\left\| f(x)-f(y) \right\|^2_2} \;.
\end{equation}
\citet{gao2021simcse} empirically showed that sentence embedding models with both lower alignment and uniformity achieve better performance in general. Similarly, we calculate the two losses on STS-B\footnote{We take STS-B pairs with a score higher than $4.0$ as $p_{\text{pos}}$ and the full STS-B as $p_{\text{data}}$. Since \citet{gao2021simcse} did not release the code for calculating these two losses, the absolute values we obtained might be different from theirs. We make sure our calculation across different models is consistent.} and results are presented in Figure \ref{fig:metric}. It shows that MCSE models achieve better alignment scores compared to SimCSE while maintaining uniformity. This analysis provides further support that visually grounding can enhance sentence representation learning by improving the alignment property of the textual embedding space.  

\subsection{Analysis}
For brevity, we take BERT-based models trained merely on caption datasets and investigate the impact of training data scales. More analysis results (sentence retrieval, cross-modal retrieval) are provided in Appendix \ref{sec:appendix_analysis}.
We limit the number of training samples to $100$, $500$, $1000$, $5000$ and $10000$, and compare their performance with the full set performance. In all of these settings, we optimize the models for same number of training steps as the full set setting. The results are shown in Figure \ref{fig:scale}. SimCSE achieves better performance than MCSE with limited samples, while MCSE starts to outperform SimCSE with the increasing data scale. We conjecture that this phenomenon can be ascribed to the progressive training of weights in multimodal projection heads. 

\section{Limitations}
Despite showing performance improvements on STS benchmarks, MCSE has its limitations as well. We take caption datasets as the source of multimodal information, while these datasets are collected and curated with non-negligible human efforts. It will have great practical value if we can properly leverage noisy image-sentence pairs or even get rid of the explicit alignments between images and sentences. Furthermore, we find that only subsets from related domains can get significant improvements while others suffer from distribution shifts. It is critical to mitigate domain gaps for learning general-purpose sentence embeddings. In addition, the definition of ``semantic similarity'' is highly task-dependent. Besides STS benchmarks, it is worth exploring the performance gap between text-only models and multimodal models on other benchmarks that can also assess the quality of sentence representations.

\section{Conclusion}
In this paper, we propose MCSE, a novel approach for sentence embedding learning that applies a multimodal contrastive objective to align sentences and corresponding images in a grounded space. Experiments show that MCSE consistently improves the performance on STS tasks. We also highlight the superiority of our method by analyzing the alignment and uniformity properties of the embedding space. The multimodal objective is generic and can be potentially incorporated into other sentence embedding methods to boost their performance.

\section*{Acknowledgements}
We thank Dingfan Chen, Fangzhou Zhai and Xiaoyu Shen for their helpful comments on the paper draft. We would also like to thank the reviewers for their valuable feedback. This work was funded by the Deutsche Forschungsgemeinschaft (DFG, German Research Foundation) – project-id 232722074 – SFB 1102. This work was also partially funded by the EU Horizon 2020 project ROXANNE under
grant number 833635 and COMPRISE grant number 3081705.

\bibliography{anthology}

\begin{thebibliography}{40}
\expandafter\ifx\csname natexlab\endcsname\relax\def\natexlab#1{#1}\fi

\bibitem[{Agirre et~al.(2015)Agirre, Banea, Cardie, Cer, Diab, Gonzalez-Agirre,
  Guo, Lopez-Gazpio, Maritxalar, Mihalcea et~al.}]{agirre2015semeval}
Eneko Agirre, Carmen Banea, Claire Cardie, Daniel Cer, Mona Diab, Aitor
  Gonzalez-Agirre, Weiwei Guo, Inigo Lopez-Gazpio, Montse Maritxalar, Rada
  Mihalcea, et~al. 2015.
\newblock \href {https://aclanthology.org/S15-2045/} {Semeval-2015 task 2:
  Semantic textual similarity, english, spanish and pilot on interpretability}.
\newblock In \emph{Proceedings of the 9th international workshop on semantic
  evaluation (SemEval 2015)}, pages 252--263.

\bibitem[{Agirre et~al.(2014)Agirre, Banea, Cardie, Cer, Diab, Gonzalez-Agirre,
  Guo, Mihalcea, Rigau, and Wiebe}]{agirre2014semeval}
Eneko Agirre, Carmen Banea, Claire Cardie, Daniel Cer, Mona Diab, Aitor
  Gonzalez-Agirre, Weiwei Guo, Rada Mihalcea, German Rigau, and Janyce Wiebe.
  2014.
\newblock \href {https://aclanthology.org/S14-2010/} {Semeval-2014 task 10:
  Multilingual semantic textual similarity}.
\newblock In \emph{Proceedings of the 8th international workshop on semantic
  evaluation (SemEval 2014)}, pages 81--91.

\bibitem[{Agirre et~al.(2016)Agirre, Banea, Cer, Diab, Gonzalez~Agirre,
  Mihalcea, Rigau~Claramunt, and Wiebe}]{agirre2016semeval}
Eneko Agirre, Carmen Banea, Daniel Cer, Mona Diab, Aitor Gonzalez~Agirre, Rada
  Mihalcea, German Rigau~Claramunt, and Janyce Wiebe. 2016.
\newblock \href {https://aclanthology.org/S16-1081/} {Semeval-2016 task 1:
  Semantic textual similarity, monolingual and cross-lingual evaluation}.
\newblock In \emph{Proceedings of the 10th International Workshop on Semantic
  Evaluation (SemEval 2016)}, pages 497--511.

\bibitem[{Agirre et~al.(2012)Agirre, Cer, Diab, and
  Gonzalez-Agirre}]{agirre2012semeval}
Eneko Agirre, Daniel Cer, Mona Diab, and Aitor Gonzalez-Agirre. 2012.
\newblock \href {https://aclanthology.org/S12-1051/} {Semeval-2012 task 6: A
  pilot on semantic textual similarity}.
\newblock In \emph{* SEM 2012: The First Joint Conference on Lexical and
  Computational Semantics--Volume 1: Proceedings of the main conference and the
  shared task, and Volume 2: Proceedings of the Sixth International Workshop on
  Semantic Evaluation (SemEval 2012)}, pages 385--393.

\bibitem[{Agirre et~al.(2013)Agirre, Cer, Diab, Gonzalez-Agirre, and
  Guo}]{agirre2013sem}
Eneko Agirre, Daniel Cer, Mona Diab, Aitor Gonzalez-Agirre, and Weiwei Guo.
  2013.
\newblock \href {https://aclanthology.org/S13-1004/} {* {SEM} 2013 shared task:
  Semantic textual similarity}.
\newblock In \emph{Second joint conference on lexical and computational
  semantics (* SEM), volume 1: proceedings of the Main conference and the
  shared task: semantic textual similarity}, pages 32--43.

\bibitem[{Bender and Koller(2020)}]{bender-koller-2020-climbing}
Emily~M. Bender and Alexander Koller. 2020.
\newblock \href {https://doi.org/10.18653/v1/2020.acl-main.463} {Climbing
  towards {NLU}: {On} meaning, form, and understanding in the age of data}.
\newblock In \emph{Proceedings of the 58th Annual Meeting of the Association
  for Computational Linguistics (ACL)}, pages 5185--5198.

\bibitem[{Bisk et~al.(2020)Bisk, Holtzman, Thomason, Andreas, Bengio, Chai,
  Lapata, Lazaridou, May, Nisnevich, Pinto, and
  Turian}]{bisk-etal-2020-experience}
Yonatan Bisk, Ari Holtzman, Jesse Thomason, Jacob Andreas, Yoshua Bengio, Joyce
  Chai, Mirella Lapata, Angeliki Lazaridou, Jonathan May, Aleksandr Nisnevich,
  Nicolas Pinto, and Joseph Turian. 2020.
\newblock \href {https://doi.org/10.18653/v1/2020.emnlp-main.703} {Experience
  grounds language}.
\newblock In \emph{Proceedings of the 2020 Conference on Empirical Methods in
  Natural Language Processing (EMNLP)}, pages 8718--8735.

\bibitem[{Bordes et~al.(2019)Bordes, Zablocki, Soulier, Piwowarski, and
  Gallinari}]{bordes2019incorporating}
Patrick Bordes, Eloi Zablocki, Laure Soulier, Benjamin Piwowarski, and Patrick
  Gallinari. 2019.
\newblock \href {https://aclanthology.org/D19-1064.pdf} {Incorporating visual
  semantics into sentence representations within a grounded space}.
\newblock In \emph{Proceedings of the 2019 Conference on Empirical Methods in
  Natural Language Processing and the 9th International Joint Conference on
  Natural Language Processing (EMNLP-IJCNLP)}, pages 696--707.

\bibitem[{Carlsson et~al.(2020)Carlsson, Gyllensten, Gogoulou, Hellqvist, and
  Sahlgren}]{carlsson2020semantic}
Fredrik Carlsson, Amaru~Cuba Gyllensten, Evangelia Gogoulou,
  Erik~Ylip{\"a}{\"a} Hellqvist, and Magnus Sahlgren. 2020.
\newblock \href {https://openreview.net/forum?id=Ov_sMNau-PF} {Semantic
  re-tuning with contrastive tension}.
\newblock In \emph{International Conference on Learning Representations
  (ICLR)}.

\bibitem[{Cer et~al.(2017)Cer, Diab, Agirre, Lopez-Gazpio, and
  Specia}]{cer2017semeval}
Daniel Cer, Mona Diab, Eneko Agirre, Inigo Lopez-Gazpio, and Lucia Specia.
  2017.
\newblock \href {https://aclanthology.org/S17-2001/} {Semeval-2017 task 1:
  Semantic textual similarity-multilingual and cross-lingual focused
  evaluation}.
\newblock In \emph{Proceedings of the 11th International Workshop on Semantic
  Evaluation (SemEval 2017)}, pages 1--14.

\bibitem[{Cer et~al.(2018)Cer, Yang, Kong, Hua, Limtiaco, St.~John, Constant,
  Guajardo-Cespedes, Yuan, Tar, Strope, and Kurzweil}]{cer-etal-2018-universal}
Daniel Cer, Yinfei Yang, Sheng-yi Kong, Nan Hua, Nicole Limtiaco, Rhomni
  St.~John, Noah Constant, Mario Guajardo-Cespedes, Steve Yuan, Chris Tar,
  Brian Strope, and Ray Kurzweil. 2018.
\newblock \href {https://doi.org/10.18653/v1/D18-2029} {Universal sentence
  encoder for {E}nglish}.
\newblock In \emph{Proceedings of the 2018 Conference on Empirical Methods in
  Natural Language Processing (EMNLP)}, pages 169--174.

\bibitem[{Chen et~al.(2020)Chen, Kornblith, Norouzi, and
  Hinton}]{chen2020simple}
Ting Chen, Simon Kornblith, Mohammad Norouzi, and Geoffrey Hinton. 2020.
\newblock \href {http://proceedings.mlr.press/v119/chen20j.html} {A simple
  framework for contrastive learning of visual representations}.
\newblock In \emph{Proceedings of the 37th International Conference on Machine
  Learning (ICML)}, pages 1597--1607.

\bibitem[{Conneau et~al.(2017)Conneau, Kiela, Schwenk, Barrault, and
  Bordes}]{conneau2017supervised}
Alexis Conneau, Douwe Kiela, Holger Schwenk, Lo{\"\i}c Barrault, and Antoine
  Bordes. 2017.
\newblock \href {https://aclanthology.org/D17-1070} {Supervised learning of
  universal sentence representations from natural language inference data}.
\newblock In \emph{Proceedings of the 2017 Conference on Empirical Methods in
  Natural Language Processing (EMNLP)}, pages 670--680.

\bibitem[{Devlin et~al.(2019)Devlin, Chang, Lee, and
  Toutanova}]{devlin2019bert}
Jacob Devlin, Ming-Wei Chang, Kenton Lee, and Kristina Toutanova. 2019.
\newblock \href {https://aclanthology.org/N19-1423} {{BERT}: Pre-training of
  deep bidirectional transformers for language understanding}.
\newblock In \emph{North {A}merican Chapter of the Association for
  Computational Linguistics: Human Language Technologies (NAACL-HIT)}, pages
  4171--4186.

\bibitem[{Gao et~al.(2021)Gao, Yao, and Chen}]{gao2021simcse}
Tianyu Gao, Xingcheng Yao, and Danqi Chen. 2021.
\newblock \href {https://arxiv.org/abs/2104.08821} {{SimCSE}: Simple
  contrastive learning of sentence embeddings}.
\newblock In \emph{Proceedings of the 2021 Conference on Empirical Methods in
  Natural Language Processing (EMNLP)}, pages 6894--6910.

\bibitem[{He et~al.(2016)He, Zhang, Ren, and Sun}]{he2016deep}
Kaiming He, Xiangyu Zhang, Shaoqing Ren, and Jian Sun. 2016.
\newblock \href
  {https://www.cv-foundation.org/openaccess/content_cvpr_2016/papers/He_Deep_Residual_Learning_CVPR_2016_paper.pdf}
  {Deep residual learning for image recognition}.
\newblock In \emph{Proceedings of the IEEE conference on computer vision and
  pattern recognition (CVPR)}, pages 770--778.

\bibitem[{Kiela et~al.(2018)Kiela, Conneau, Jabri, and
  Nickel}]{kiela2018learning}
Douwe Kiela, Alexis Conneau, Allan Jabri, and Maximilian Nickel. 2018.
\newblock \href {https://aclanthology.org/N18-1038} {Learning visually grounded
  sentence representations}.
\newblock In \emph{Proceedings of the 2018 Conference of the North American
  Chapter of the Association for Computational Linguistics: Human Language
  Technologies (NAACL-HIT)}, pages 408--418.

\bibitem[{Kim et~al.(2021)Kim, Yoo, and Lee}]{kim-etal-2021-self}
Taeuk Kim, Kang~Min Yoo, and Sang-goo Lee. 2021.
\newblock \href {https://doi.org/10.18653/v1/2021.acl-long.197} {Self-guided
  contrastive learning for {BERT} sentence representations}.
\newblock In \emph{Proceedings of the 59th Annual Meeting of the Association
  for Computational Linguistics and the 11th International Joint Conference on
  Natural Language Processing (ACL-IJCNLP)}, pages 2528--2540.

\bibitem[{Kiros et~al.(2015)Kiros, Zhu, Salakhutdinov, Zemel, Urtasun,
  Torralba, and Fidler}]{kiros2015skip}
Ryan Kiros, Yukun Zhu, Russ~R Salakhutdinov, Richard Zemel, Raquel Urtasun,
  Antonio Torralba, and Sanja Fidler. 2015.
\newblock \href
  {https://proceedings.neurips.cc/paper/2015/file/f442d33fa06832082290ad8544a8da27-Paper.pdf}
  {Skip-thought vectors}.
\newblock In \emph{Advances in neural information processing systems
  (NeurIPS)}, pages 3294--3302.

\bibitem[{Lazaridou et~al.(2015)Lazaridou, Baroni
  et~al.}]{lazaridou2015combining}
Angeliki Lazaridou, Marco Baroni, et~al. 2015.
\newblock \href {https://aclanthology.org/N15-1016} {Combining language and
  vision with a multimodal skip-gram model}.
\newblock In \emph{Proceedings of the 2015 Conference of the North American
  Chapter of the Association for Computational Linguistics: Human Language
  Technologies (NAACL-HLT)}, pages 153--163.

\bibitem[{Li et~al.(2020)Li, Zhou, He, Wang, Yang, and
  Li}]{li-etal-2020-sentence}
Bohan Li, Hao Zhou, Junxian He, Mingxuan Wang, Yiming Yang, and Lei Li. 2020.
\newblock \href {https://doi.org/10.18653/v1/2020.emnlp-main.733} {On the
  sentence embeddings from pre-trained language models}.
\newblock In \emph{Proceedings of the 2020 Conference on Empirical Methods in
  Natural Language Processing (EMNLP)}, pages 9119--9130.

\bibitem[{Lin et~al.(2014)Lin, Maire, Belongie, Hays, Perona, Ramanan,
  Doll{\'a}r, and Zitnick}]{lin2014coco}
Tsung-Yi Lin, Michael Maire, Serge Belongie, James Hays, Pietro Perona, Deva
  Ramanan, Piotr Doll{\'a}r, and C.~Lawrence Zitnick. 2014.
\newblock \href
  {https://link.springer.com/chapter/10.1007/978-3-319-10602-1_48} {Microsoft
  coco: Common objects in context}.
\newblock In \emph{Proceedings of the European Conference on Computer Vision
  (ECCV)}, pages 740--755.

\bibitem[{Liu et~al.(2021)Liu, Vuli{\'c}, Korhonen, and
  Collier}]{liu-etal-2021-fast}
Fangyu Liu, Ivan Vuli{\'c}, Anna Korhonen, and Nigel Collier. 2021.
\newblock \href {https://doi.org/10.18653/v1/2021.emnlp-main.109} {Fast,
  effective, and self-supervised: Transforming masked language models into
  universal lexical and sentence encoders}.
\newblock In \emph{Proceedings of the 2021 Conference on Empirical Methods in
  Natural Language Processing (EMNLP)}, pages 1442--1459.

\bibitem[{Liu et~al.(2019)Liu, Ott, Goyal, Du, Joshi, Chen, Levy, Lewis,
  Zettlemoyer, and Stoyanov}]{liu2019roberta}
Yinhan Liu, Myle Ott, Naman Goyal, Jingfei Du, Mandar Joshi, Danqi Chen, Omer
  Levy, Mike Lewis, Luke Zettlemoyer, and Veselin Stoyanov. 2019.
\newblock \href {https://arxiv.org/abs/1907.11692} {Roberta: A robustly
  optimized bert pretraining approach}.
\newblock \emph{arXiv preprint arXiv:1907.11692}.

\bibitem[{Marelli et~al.(2014)Marelli, Menini, Baroni, Bentivogli, Bernardi,
  Zamparelli et~al.}]{marelli2014sick}
Marco Marelli, Stefano Menini, Marco Baroni, Luisa Bentivogli, Raffaella
  Bernardi, Roberto Zamparelli, et~al. 2014.
\newblock \href
  {http://www.lrec-conf.org/proceedings/lrec2014/pdf/363_Paper.pdf} {A sick
  cure for the evaluation of compositional distributional semantic models.}
\newblock In \emph{International Conference on Language Resources and
  Evaluation (LREC)}, pages 216--223.

\bibitem[{Pennington et~al.(2014)Pennington, Socher, and
  Manning}]{pennington2014glove}
Jeffrey Pennington, Richard Socher, and Christopher~D Manning. 2014.
\newblock \href {https://aclanthology.org/D14-1162.pdf} {Glove: Global vectors
  for word representation}.
\newblock In \emph{Proceedings of the 2014 conference on empirical methods in
  natural language processing (EMNLP)}, pages 1532--1543.

\bibitem[{Radford et~al.(2021)Radford, Kim, Hallacy, Ramesh, Goh, Agarwal,
  Sastry, Askell, Mishkin, Clark, Krueger, and Sutskever}]{radford2021clip}
Alec Radford, Jong~Wook Kim, Chris Hallacy, Aditya Ramesh, Gabriel Goh,
  Sandhini Agarwal, Girish Sastry, Amanda Askell, Pamela Mishkin, Jack Clark,
  Gretchen Krueger, and Ilya Sutskever. 2021.
\newblock \href {http://proceedings.mlr.press/v139/radford21a.html} {Learning
  transferable visual models from natural language supervision}.
\newblock In \emph{Proceedings of the 38th International Conference on Machine
  Learning (ICML)}, pages 8748--8763.

\bibitem[{Reimers and Gurevych(2019)}]{reimers-gurevych-2019-sentence}
Nils Reimers and Iryna Gurevych. 2019.
\newblock \href {https://doi.org/10.18653/v1/D19-1410} {Sentence-{BERT}:
  Sentence embeddings using {S}iamese {BERT}-networks}.
\newblock In \emph{Proceedings of the 2019 Conference on Empirical Methods in
  Natural Language Processing and the 9th International Joint Conference on
  Natural Language Processing (EMNLP-IJCNLP)}, pages 3982--3992.

\bibitem[{Su et~al.(2021)Su, Cao, Liu, and Ou}]{su2021whitening}
Jianlin Su, Jiarun Cao, Weijie Liu, and Yangyiwen Ou. 2021.
\newblock \href {https://arxiv.org/pdf/2103.15316.pdf} {Whitening sentence
  representations for better semantics and faster retrieval}.
\newblock \emph{arXiv preprint arXiv:2103.15316}.

\bibitem[{Tan and Bansal(2020)}]{tan2020vokenization}
Hao Tan and Mohit Bansal. 2020.
\newblock \href {https://aclanthology.org/2020.emnlp-main.162.pdf}
  {Vokenization: Improving language understanding via contextualized,
  visually-grounded supervision}.
\newblock In \emph{Proceedings of the 2020 Conference on Empirical Methods in
  Natural Language Processing (EMNLP)}, pages 2066--2080.

\bibitem[{Tang et~al.(2021)Tang, Cho, Tan, and Bansal}]{tang2021vidlankd}
Zineng Tang, Jaemin Cho, Hao Tan, and Mohit Bansal. 2021.
\newblock \href
  {https://papers.nips.cc/paper/2021/file/ccdf3864e2fa9089f9eca4fc7a48ea0a-Paper.pdf}
  {Vidlankd: Improving language understanding via video-distilled knowledge
  transfer}.
\newblock In \emph{Advances in Neural Information Processing Systems
  (NeurIPS)}, pages 24468--24481.

\bibitem[{Wang and Isola(2020)}]{wang2020understanding}
Tongzhou Wang and Phillip Isola. 2020.
\newblock \href {http://proceedings.mlr.press/v119/wang20k/wang20k.pdf}
  {Understanding contrastive representation learning through alignment and
  uniformity on the hypersphere}.
\newblock In \emph{Proceedings of the 37th International Conference on Machine
  Learning (ICML)}, pages 9929--9939.

\bibitem[{Wieting et~al.(2020)Wieting, Neubig, and
  Berg-Kirkpatrick}]{wieting2020bilingual}
John Wieting, Graham Neubig, and Taylor Berg-Kirkpatrick. 2020.
\newblock \href {https://aclanthology.org/2020.emnlp-main.122.pdf} {A bilingual
  generative transformer for semantic sentence embedding}.
\newblock In \emph{Proceedings of the 2020 Conference on Empirical Methods in
  Natural Language Processing (EMNLP)}, pages 1581--1594.

\bibitem[{Wolf et~al.(2020)Wolf, Debut, Sanh, Chaumond, Delangue, Moi, Cistac,
  Rault, Louf, Funtowicz, Davison, Shleifer, von Platen, Ma, Jernite, Plu, Xu,
  Scao, Gugger, Drame, Lhoest, and Rush}]{wolf-etal-2020-transformers}
Thomas Wolf, Lysandre Debut, Victor Sanh, Julien Chaumond, Clement Delangue,
  Anthony Moi, Pierric Cistac, Tim Rault, Rémi Louf, Morgan Funtowicz, Joe
  Davison, Sam Shleifer, Patrick von Platen, Clara Ma, Yacine Jernite, Julien
  Plu, Canwen Xu, Teven~Le Scao, Sylvain Gugger, Mariama Drame, Quentin Lhoest,
  and Alexander~M. Rush. 2020.
\newblock \href {https://www.aclweb.org/anthology/2020.emnlp-demos.6}
  {Transformers: State-of-the-art natural language processing}.
\newblock In \emph{Proceedings of the 2020 Conference on Empirical Methods in
  Natural Language Processing (EMNLP)}, pages 38--45.

\bibitem[{Yan et~al.(2021)Yan, Li, Wang, Zhang, Wu, and
  Xu}]{yan-etal-2021-consert}
Yuanmeng Yan, Rumei Li, Sirui Wang, Fuzheng Zhang, Wei Wu, and Weiran Xu. 2021.
\newblock \href {https://doi.org/10.18653/v1/2021.acl-long.393} {{C}on{SERT}: A
  contrastive framework for self-supervised sentence representation transfer}.
\newblock In \emph{Proceedings of the 59th Annual Meeting of the Association
  for Computational Linguistics and the 11th International Joint Conference on
  Natural Language Processing (EMNLP-IJCNLP)}, pages 5065--5075.

\bibitem[{Young et~al.(2014)Young, Lai, Hodosh, and
  Hockenmaier}]{young2014flickr}
Peter Young, Alice Lai, Micah Hodosh, and Julia Hockenmaier. 2014.
\newblock \href {https://doi.org/10.1162/tacl_a_00166} {{From image
  descriptions to visual denotations: New similarity metrics for semantic
  inference over event descriptions}}.
\newblock \emph{Transactions of the Association for Computational Linguistics
  (TACL)}, pages 67--78.

\bibitem[{Zablocki et~al.(2018)Zablocki, Piwowarski, Soulier, and
  Gallinari}]{zablocki2018learning}
Eloi Zablocki, Benjamin Piwowarski, Laure Soulier, and Patrick Gallinari. 2018.
\newblock \href {https://ojs.aaai.org/index.php/AAAI/article/view/11939}
  {Learning multi-modal word representation grounded in visual context}.
\newblock In \emph{Proceedings of the AAAI Conference on Artificial
  Intelligence}.

\bibitem[{Zhang et~al.(2020)Zhang, He, Liu, Lim, and
  Bing}]{zhang-etal-2020-unsupervised}
Yan Zhang, Ruidan He, Zuozhu Liu, Kwan~Hui Lim, and Lidong Bing. 2020.
\newblock \href {https://doi.org/10.18653/v1/2020.emnlp-main.124} {An
  unsupervised sentence embedding method by mutual information maximization}.
\newblock In \emph{Proceedings of the 2020 Conference on Empirical Methods in
  Natural Language Processing (EMNLP)}, pages 1601--1610.

\bibitem[{Zhang et~al.(2019)Zhang, Chen, Wang, Utiyama, Sumita, Li, and
  Zhao}]{zhang2019neural}
Zhuosheng Zhang, Kehai Chen, Rui Wang, Masao Utiyama, Eiichiro Sumita, Zuchao
  Li, and Hai Zhao. 2019.
\newblock \href {https://openreview.net/forum?id=Byl8hhNYPS} {Neural machine
  translation with universal visual representation}.
\newblock In \emph{International Conference on Learning Representations
  (ICLR)}.

\bibitem[{Zhao and Titov(2020)}]{zhao2020visually}
Yanpeng Zhao and Ivan Titov. 2020.
\newblock \href {https://aclanthology.org/2020.emnlp-main.354.pdf} {Visually
  grounded compound pcfgs}.
\newblock In \emph{Proceedings of the 2020 Conference on Empirical Methods in
  Natural Language Processing (EMNLP)}, pages 4369--4379.

\end{thebibliography}
\bibliographystyle{acl_natbib}

\clearpage
\appendix

\section{Implementation Details}
\label{sec:appendix_implementation}
\textbf{Language Encoder} 
Our implementation is based on the Hugging Face Transformers library\footnote{\href{https://github.com/huggingface/transformers}{https://github.com/huggingface/transformers}}~\citep{wolf-etal-2020-transformers}. We start from the checkpoints of \texttt{bert-base-uncased} and \texttt{roberta-base}, and fine-tune the pre-trained models using a contrastive objective function. We use the 768-dimensional \texttt{[CLS]} token outputs before the MLP pooler layer as sentence embeddings for evaluation. 

\noindent
\textbf{Image Encoder} 
We use ResNet-50 and extract 2048-dimensional feature vectors at the last layer. The image encoder is not fine-tuned.\footnote{In our preliminary results, fine-tuning the image encoder does not have a significant impact on the STS performance.}

\noindent
\textbf{Projection Heads} 
We use distinct projection heads for different modalities and objectives. All of them are implemented by single-layer MLPs with Tanh activation. We map sentence embeddings to a 768-dimensional space before calculating the textual objective. We map both sentence embeddings and image feature vectors to a 256-dimensional shared space, and normalize them before calculating the multimodal objective.   

\noindent
\textbf{Parameter Settings} We explore $5$ training settings in the paper: \{\textit{wiki}, \textit{wiki+flickr}, \textit{wiki+coco}, \textit{flickr}, \textit{coco}\}. For \textit{wiki+flickr} and \textit{wiki+coco}, we sample mini-batches from either Wiki1M or the caption dataset in proportion to their data size. We adopt most of the parameter settings suggested by \citet{gao2021simcse}. Moreover, temperature parameters $\tau$ and $\tau'$ are set to $0.05$, and other hyperparameters are reported in Table \ref{tab:hyperparameter}. We use the dev set of STS-B to tune the trade-off parameter $\lambda$ and ablation studies are shown in Table \ref{tab:lambda}. We evaluate models every $125$ training steps on STS-B dev set and keep the best checkpoint for final evaluation.

\begin{table}[h]
 \begin{center}
 \scalebox{0.71}{
  \begin{tabular}{l|ccccc}
  \toprule
   \textbf{settings:} & \textit{wiki} & \textit{wiki+flickr} & \textit{wiki+coco} & \textit{flickr} & \textit{coco} \\
  \midrule
  \multicolumn{6}{c}{BERT} \\
  \midrule
  learning rate & \multicolumn{5}{c}{3e-5}\\
  batch size & \multicolumn{5}{c}{64} \\
  $\lambda$ & -- & 0.01 & 0.01 & 0.05 & 0.05 \\
  epochs & 3 & 3& 3&6&3 \\
  \midrule
  \multicolumn{6}{c}{RoBERTa} \\
  \midrule
  learning rate & \multicolumn{5}{c}{1e-5}\\
  batch size & \multicolumn{5}{c}{128} \\
  $\lambda$ &  -- & 0.01 & 0.01 &0.01 & 0.01 \\
  epochs & 3 & 3& 3&6&3 \\
  \bottomrule
  \end{tabular}}
  \caption{The hyperparameters used for different training settings and pre-trained encoders.}
  \label{tab:hyperparameter}
  \end{center}
\end{table}

\begin{table}[h]
\begin{center}
 \scalebox{0.75}{
  \begin{tabular}{lccccc}
  \toprule
  $\lambda$ & 0.001 & 0.01 & 0.05 & 0.1 & 0.5 \\
  \midrule
  MCSE-BERT & 78.38 & 79.95 & \textbf{80.41} & 80.35 & 80.01 \\
  MCSE-RoBERTa & 80.60 & \textbf{81.48} & 81.08 & 80.73 & 79.85 \\
  \bottomrule
  \end{tabular}}
  \caption{STS-B performance of MCSE models trained on Flickr30k with different trade-off parameters.}
  \label{tab:lambda}
 \end{center}
\end{table}

\section{More Results}
\label{sec:appendix_b}

\subsection{Improvements on Different Subsets}
\label{sec:appendix_subset}
\begin{figure*}[t]
    \centering
    \begin{subfigure}{1.0 \textwidth}
        \centering
        \includegraphics[scale=0.66]{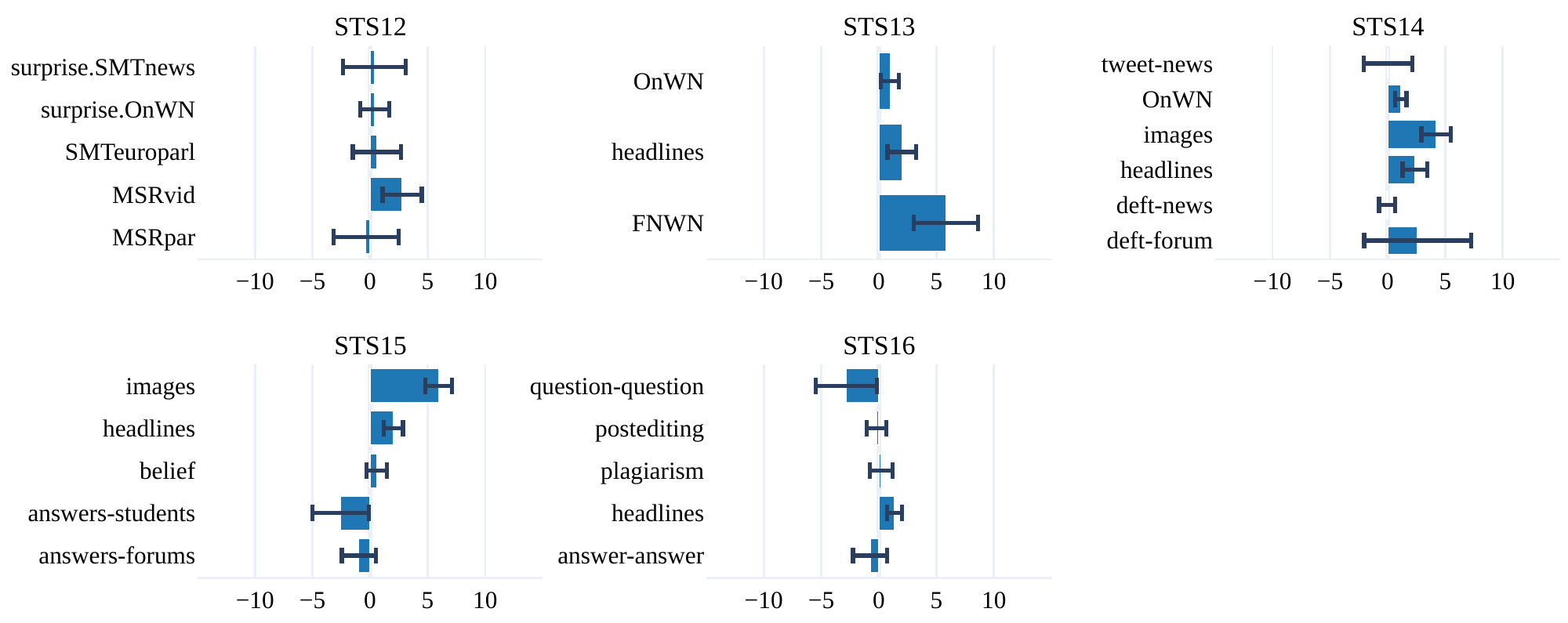} 
        \caption{training data: \textit{wiki+flickr} }
    \end{subfigure}
    \newline
    \begin{subfigure}{1.0 \textwidth}
        \centering
        \includegraphics[scale=0.7]{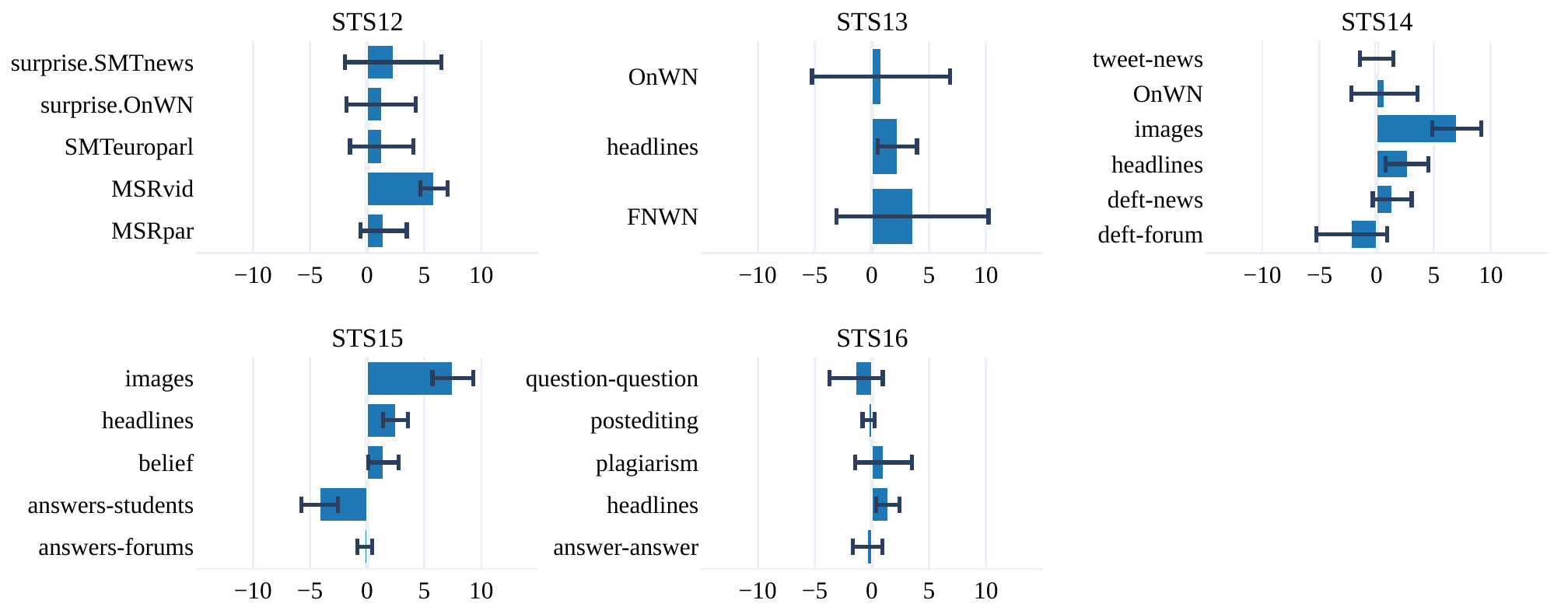} 
        \caption{training data: \textit{wiki+coco} }
    \end{subfigure}
    \caption{The Spearman's correlation improvements over different subsets. }
    \label{fig:subset}
\end{figure*}

To delve into the performance gap between \texttt{MCSE-BERT} and \texttt{SimCSE-BERT},  we calculate the Spearman's correlation for different subsets of each year's STS challenge separately. The improvements of MCSE over SimCSE  are shown in Figure \ref{fig:subset}. In STS12, "MSRvid" subset achieves the largest improvement, which is a corpus of video descriptions. "Image" subsets in STS14 and STS15 also get considerable improvements. Meanwhile, the performance of "answers-students" subset in STS15 drops extensively, and none of the subsets in STS16 get noticeable improvement by MCSE. The results indicate that the subsets benefit to different degrees from the visually grounding because of domain discrepancy.

\subsection{Ablation Study}
\label{sec:appendix_ablation}
\paragraph{CLIP as Image Encoder}
We use CLIP~\citep{radford2021clip} as an alternative image encoder. The implementation is based on the Sentence Transformer library\footnote{\href{https://github.com/UKPLab/sentence-transformers}{https://github.com/UKPLab/sentence-transformers}}~\citep{reimers-gurevych-2019-sentence} and we use the checkpoint \texttt{clip-ViT-B-32} to extract 512-dimensional feature vectors. As shown in Table \ref{tab:clip}, different image encoders lead to very similar results, thus we use ResNet as the default image encoder.

\paragraph{Combining Wiki1M, Flickr30k and MS-COCO} 
We adopt the same parameter setting as \textit{wiki+flickr} and \textit{wiki+coco}, and train models on the combination of Wiki1M, Flickr30k, and MS-COCO. As shown in Table \ref{tab:all}, MCSE models achieve $1.9$ point and $2.6$ point improvements when using BERT and RoBERTa, respectively.

\begin{table}[ht]
 \begin{center}
 \scalebox{0.75}{
  \begin{tabular}{lc}
    \toprule
    \multirow{2.5}{*}{\textbf{Model}} & \textbf{Trained on}\\
     & \textit{wiki+flickr+coco}\\
    \midrule
    \midrule
    SimCSE-BERT & 74.3$_{\pm 1.0}$ \\
    \rowcolor{Color} 
    MCSE-BERT & \textbf{76.2}$_{\pm 0.3}$ \\
    \midrule
    SimCSE-RoBERTa & 75.3$_{\pm 0.7}$ \\
    \rowcolor{Color} 
    MCSE-RoBERTa & \textbf{77.9}$_{\pm 0.6}$ \\
    \bottomrule
  \end{tabular}}\\
  \caption{ Comparison of the average Spearman's correlation of $7$ STS tasks. We report the means and standard deviations over $5$ random seeds.}
  \label{tab:all}
  \end{center}
\end{table}

\begin{table}[h]
 \begin{center}
 \scalebox{0.75}{
  \begin{tabular}{lcccc}
    \toprule
    \multirow{2.5}{*}{\textbf{Model}} & \multicolumn{2}{c}{\textbf{\text{image } $\rightarrow$ \text{text}}} &
    \multicolumn{2}{c}{\textbf{\text{text } $\rightarrow$ \text{image}}}\\
    \cmidrule{2-5}
     & R@1 & R@5 & R@1 & R@5 \\
    \midrule
    \midrule
    MCSE-BERT\textsubscript{\textit{wiki+flickr}} & 16.7&	43.5&	22.5&	50.4  \\
    MCSE-BERT\textsubscript{\textit{flickr}} & 20.4&	50.2&	23.8&	52.5  \\
    \midrule
    MCSE-BERT\textsubscript{\textit{wiki+coco}} & 8.8&	26.6&	10.9&	31.2  \\
    MCSE-BERT\textsubscript{\textit{coco}} &  8.2&	25.2&	9.0&	27.1  \\
  \bottomrule
  \end{tabular}} \\
  \caption{Multimodal retrieval results on Flickr30k test set (1k) and MS-COCO minival set (5k). }
  \label{tab:cross}
  \end{center}
\end{table}

\subsection{Analysis}
\label{sec:appendix_analysis}
\paragraph{Sentence Retrieval}
We take BERT-based models trained on the Flickr30k train set (same seed) and conduct a sentence retrieval experiment on Flickr30k test set. Given an input sentence, the nearest neighbor will be retrieved based on cosine similarity. Some retrieval examples are shown in Table \ref{tab:more_example}. We observe that (1) SimCSE is prone to retrieving sentences with similar syntax, while MCSE can retrieve sentences that vary in syntax and share semantics. Examples: Q1, Q3, Q6. (2) MCSE is better at recognizing similar event scenes and capturing the number of entities. Examples: Q2, Q4, Q5.

\paragraph{Cross-Modal Retrieval}
We take BERT-based models (same seed) and conduct cross-modal retrieval experiments. We use the metric Recall@K, which is calculated based on if the ground truth of the query image or caption appears in the top-K retrieved captions or images. As results in Table \ref{tab:cross} show, MCSE models also achieve a decent level of retrieval performance as a by-product of multimodal contrastive learning.

\begin{table*}[ht]
 \begin{center}
 \scalebox{0.75}{
  \begin{tabular}{clccccccc|c}
    \toprule
    &\textbf{Model} & \textbf{STS12} & \textbf{STS13} & \textbf{STS14} & \textbf{STS15} & \textbf{STS16} & \textbf{STS-B} & \textbf{SICK-R} & \textbf{Avg.$\uparrow$} \\
    \midrule
    \parbox[t]{2mm}{\multirow{6}{*}{\rotatebox[origin=c]{90}{\textit{flickr}}}} 
    &SimCSE-BERT & 
    62.1$_{\pm 0.5}$ &	
    73.8$_{\pm 0.9}$ &
    64.2$_{\pm 0.6}$ &
    74.2$_{\pm 0.8}$ &
    \textbf{74.8}$^*_{\pm 0.6}$ &
    67.1$_{\pm 1.1}$ &
    65.4$_{\pm 1.1}$ &
    68.8$_{\pm 0.7}$  \\
    &MCSE-ResNet-BERT & 
    \textbf{63.6}$^*_{\pm 0.7}$ &
    \textbf{74.0}$_{\pm 0.9}$ & 
    65.5$_{\pm  1.1}$  &
    75.5$_{\pm  0.2}$ &
    71.6$_{\pm0.4}$ &	
    74.0$_{\pm  0.4}$ &
    69.8$_{\pm  0.3}$ &
    70.6$_{\pm  0.5}$  \\
    &MCSE-CLIP-BERT & 
    63.1$_{\pm 0.7}$ &  
    73.9$_{\pm 1.0}$ &
    \textbf{65.8}$^*_{\pm 0.9}$ &
    \textbf{76.0}$^*_{\pm 0.7}$ &
    70.7$_{\pm 0.3}$ &
    \textbf{74.9}$^*_{\pm 0.5}$ &
    \textbf{70.7}$^*_{\pm 0.3}$ &
    \textbf{70.7}$^*_{\pm 0.2}$ \\
    \cmidrule{2-10}
    &SimCSE-RoBERTa & 
    66.6$_{\pm 0.5}$  &
    78.3$_{\pm 0.5}$ &
    69.7$_{\pm 0.6}$ &
    77.7$_{\pm 0.5}$  &
    \textbf{76.3}$^*_{\pm 0.5}$  &
    75.8$_{\pm 0.3}$  &
    66.2$_{\pm 0.4}$  &
    72.9$_{\pm 0.3}$ \\
    &MCSE-ResNet-RoBERTa &  
    \textbf{67.6}$^*_{\pm 0.5}$ &
    \textbf{78.8}$_{\pm 0.4}$ &
    \textbf{70.1}$_{\pm 0.3}$ &
    78.5$_{\pm 0.2}$ &
    75.4$_{\pm 0.5}$ &
    \textbf{77.4}$^*_{\pm 0.3}$ &
    68.6$_{\pm 0.3}$ &
    \textbf{73.8}$^*_{\pm 0.2}$ \\
    &MCSE-CLIP-RoBERTa & 
    67.0$_{\pm 0.5}$ &
    78.6$_{\pm 0.4}$ &
    69.8$_{\pm 0.5}$ &
    \textbf{78.7}$^*_{\pm 0.8}$ &
    74.9$_{\pm 0.5}$ &
    \textbf{77.4}$^*_{\pm 0.4}$ &
    \textbf{69.5}$^*_{\pm 0.5}$ &
    73.7$_{\pm 0.2}$\\
    \midrule
    \parbox[t]{2mm}{\multirow{6}{*}{\rotatebox[origin=c]{90}{\textit{coco}}}}
    &SimCSE-BERT &
    59.3$_{\pm 0.9}$ & 
    73.0$_{\pm 1.2}$ &
    62.7$_{\pm 0.6}$ &
    74.7$_{\pm 0.7}$ &
    \textbf{74.4}$^*_{\pm 0.4}$ &
    65.3$_{\pm 0.7}$ &
    65.4$_{\pm 0.5}$ &
    67.8$_{\pm 0.4}$ \\
    &MCSE-ResNet-BERT &
    \textbf{64.9}$^*_{\pm 0.5}$ &
    \textbf{74.8}$^*_{\pm 0.9}$ &
    \textbf{68.1}$^*_{\pm 0.6}$ &
    \textbf{76.8}$^*_{\pm 0.6}$ &
    72.7$_{\pm 0.8}$ &
    \textbf{74.5}$^*_{\pm 0.4}$ &
    69.7$_{\pm 0.4}$ &
    \textbf{71.6}$^*_{\pm 0.2}$  \\
    &MCSE-CLIP-BERT &  
    64.8$_{\pm 0.6}$ &
    74.1$_{\pm 0.6}$ &
    68.0$_{\pm 0.2}$ &
    76.2$_{\pm 0.5}$ &
    71.6$_{\pm 0.4}$ &
    \textbf{74.5}$^*_{\pm 0.3}$ &
    \textbf{70.3}$^*_{\pm 0.6}$ &
    71.4$_{\pm 0.1}$  \\
    \cmidrule{2-10}
    & SimCSE-RoBERTa &   
    64.7$_{\pm 0.6}$ &
    79.2$_{\pm 0.4}$ &
    70.2$_{\pm 0.4}$ &
    79.0$_{\pm 0.6}$ &
    \textbf{78.2}$_{\pm 0.5}$ &
    73.8$_{\pm 0.5}$ &
    64.6$_{\pm 0.3}$ &
    72.8$_{\pm 0.3}$  \\
    &MCSE-ResNet-RoBERTa &  
    \textbf{67.0}$^*_{\pm 0.8}$ &
    \textbf{79.4}$_{\pm 0.4}$ &
    \textbf{70.9}$^*_{\pm 0.4}$ &
    \textbf{80.0}$^*_{\pm 0.4}$ &
    77.8$_{\pm 0.5}$ &
    \textbf{76.9}$^*_{\pm 0.4}$ &
    67.9$_{\pm 0.7}$ &
    \textbf{74.3}$^*_{\pm 0.3}$  \\
    &MCSE-CLIP-RoBERTa &  
    66.0$_{\pm 1.0}$ &
    79.0$_{\pm 0.7}$ &
    70.6$_{\pm 0.6}$ &
    \textbf{80.0}$^*_{\pm 0.8}$ &
    77.6$_{\pm 0.5}$ &
    76.5$_{\pm 0.4}$ &
    \textbf{68.4}$^*_{\pm 0.8}$ &
    74.0$_{\pm 0.2}$  \\
    \bottomrule
 \end{tabular}}\\
   \vspace{1mm}
      \begin{small}
      $\ast$: difference between SimCSE and MCSE (ResNet/CLIP) is significant at $\alpha=0.05$ according to an independent t-test.
     \end{small}
  \caption{Performance comparison on STS tasks. STS-B: STS Benchmark, SICK-R: SICK-Relatedness, Avg.: average across $7$ tasks. Models are trained with $5$ random seeds and we report means and standard deviations.}
  \label{tab:clip}
  \end{center}
\end{table*}

\begin{table*}[h]
 \begin{center}
 \scalebox{0.85}{
  \begin{tabular}{ll}
  \toprule
  \textbf{Model} & \textbf{Result} \\
  \midrule
  \multicolumn{2}{l}{\textbf{Query 1}: A young girl is washing her teddy bear in the kitchen sink. } \\
  \midrule
  SimCSE: & A middle-aged woman is vacuuming her kitchen floor with a canister vac.   \\
  MCSE: & A young girl, blond and wearing a polka-dot shirt, washes a stuffed animal in a vanity sink. \\
  \midrule
  \multicolumn{2}{l}{\textbf{Query 2}: Three chefs , wearing white hats and black aprons , are preparing food in a crowded kitchen.} \\
  \midrule
  SimCSE: &  Numerous workers with blue shirts and white aprons are preparing fish for sale.  \\
  MCSE: & Three men are preparing food in a kitchen setting. \\
  \midrule
  \multicolumn{2}{l}{\textbf{Query 3}: A couple kisses in a shady walkway. } \\
  \midrule
  SimCSE: &   A couple strolls down a path near benches and water. \\
  MCSE: & Couple kissing outside on street. \\
  \midrule
  \multicolumn{2}{l}{\textbf{Query 4}: A man is standing on the streets taking photographs. } \\
  \midrule
  SimCSE: &   People run a marathon on a city street with a crowd watching.  \\
  MCSE: &   A guy wearing a white shirt is taking a picture. \\
  \midrule
  \multicolumn{2}{l}{\textbf{Query 5}: Two boys are playing in pool filled with sparkling blue water. } \\
  \midrule
  SimCSE: &   A little girl is swimming under the crystal blue water.  \\
  MCSE: & Two children are swimming in a pool. \\
  \midrule
  \multicolumn{2}{l}{\textbf{Query 6}: An old man sitting on a bench staring at the ocean.} \\
  SimCSE: &  A man sitting on a bench by the ocean.  \\
  MCSE: & An old man sits on a bench overlooking the water. \\
  \bottomrule
  \end{tabular}}
  \caption{Retrieved examples from Flickr30k test set.}
  \label{tab:more_example}
  \end{center}
\end{table*}

\end{document}